\documentclass{llncs}
\usepackage{url}

\usepackage{graphicx}
\usepackage[backend=bibtex,url=false,eprint=false,isbn=false, doi=false]{biblatex}
\newcommand\blfootnote[1]{
  \begingroup
  \renewcommand\thefootnote{}\footnote{#1}
  \addtocounter{footnote}{-1}
  \endgroup
}

\begin{document}
\mainmatter
\title{Co-optimising Robot Morphology and Controller in a Simulated Open-ended Environment}
\titlerunning{Co-optimising Morphology and Controller in an Open-ended Environment}

\author{Emma Hjellbrekke Stensby\inst{1} \and Kai Olav Ellefsen\inst{1} \and Kyrre Glette\inst{1,2}}

\institute{Department of Informatics, University of Oslo, Oslo, Norway \and RITMO, University of Oslo, Oslo, Norway}

\maketitle

\begin{abstract}
Designing\blfootnote{The final publication is available at Springer via \url{https://doi.org/10.1007/978-3-030-72699-7_3}} robots by hand can be costly and time consuming, especially if the robots have to be created with novel materials, or be robust to internal or external changes. In order to create robots automatically, without the need for human intervention, it is necessary to optimise both the behaviour and the body design of the robot. However, when co-optimising the morphology and controller of a locomoting agent the morphology tends to converge prematurely, reaching a local optimum. Approaches such as explicit protection of morphological innovation have been used to reduce this problem, but it might also be possible to increase exploration of morphologies using a more indirect approach.
We explore how changing the environment, where the agent locomotes, affects the convergence of morphologies. The agents' morphologies and controllers are co-optimised, while the environments the agents locomote in are evolved open-endedly with the Paired Open-Ended Trailblazer (POET). We compare the diversity, fitness and robustness of agents evolving in environments generated by POET to agents evolved in handcrafted curricula of environments. 
Our agents each contain of a population of individuals being evolved with a genetic algorithm. This population is called the agent-population. We show that agent-populations evolving in open-endedly evolving environments exhibit larger morphological diversity than agent-populations evolving in hand crafted curricula of environments. POET proved capable of creating a curriculum of environments which encouraged both diversity and quality in the populations. This suggests that POET may be capable of reducing premature convergence in co-optimisation of morphology and controllers.

\end{abstract}

\section{Introduction}
Finding a morphology and controller for a robot, that allows the robot to efficiently complete its task, is a difficult endeavour. Creating and programming robots by hand is feasible when the robots' working environment is predictable, such as in a factory or warehouse. However, it becomes almost impossible when the robots are acting outside in a constantly changing world. When a robot needs to adapt to a variety of new environments, evolutionary algorithms can be used to automatically optimise both morphology and controllers \cite{Auerbach,miras2020environmental,dyret3}.

When simultaneously evolving the controller and morphology of a robot the controller has a tendency to specialise in the current morphology \cite{diff}. If the morphology is changed the controller might no longer work. The morphology and controller are strongly connected, and when the morphology changes it is like the interface between them has been scrambled. This connection between controller and morphology can cause the morphological search to stagnate: When the controller has adapted to the morphology, the morphology may stop changing, as changes will be likely to lower the individual's fitness. Approaches to tackle this problem include directly or indirectly protecting individuals that recently experienced change in their morphology \cite{scale}, or optimising for morphological novelty in addition to fitness in a multi objective search \cite{NSLC}. However, we believe it might be possible to increase the exploration of morphologies by evolving the agents in changing environments.

\begin{figure}
\includegraphics[width=340pt]{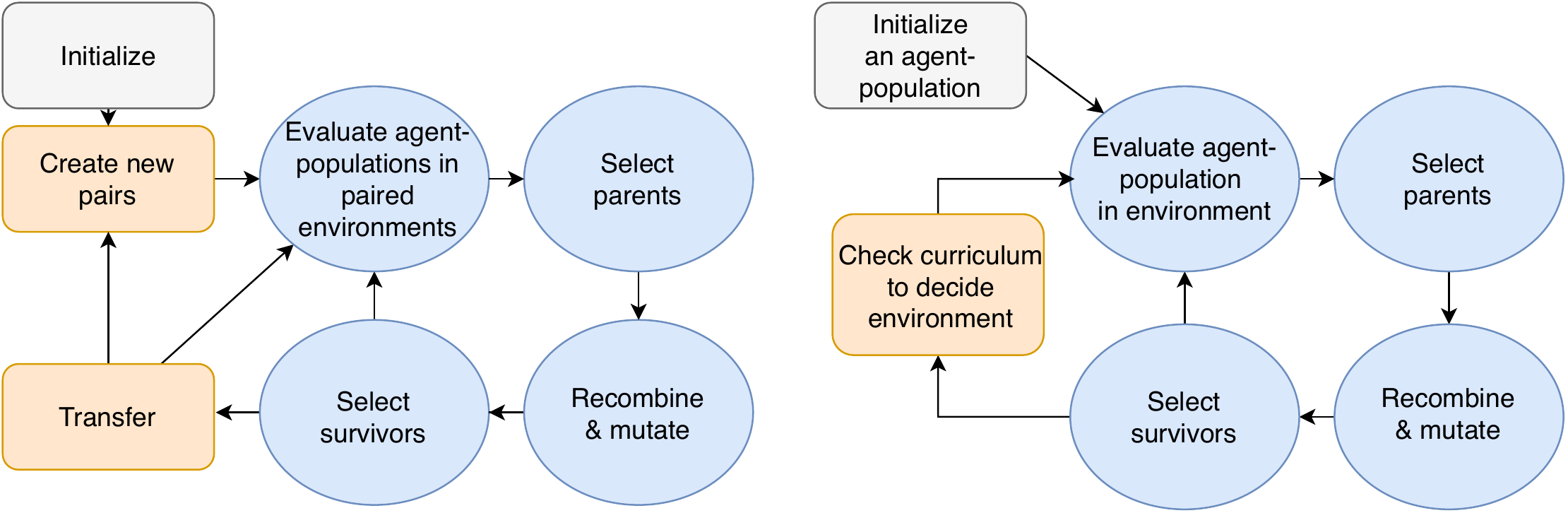}
\caption{\textbf{Left: } Illustration of our genetic algorithm when used with POET. \textbf{Right: } Illustration of our genetic algorithm when used with a curriculum of environments. (Note that in this context the term agent-population refers to a population of 192 individuals)}
\label{fig:poet_ga_loop}
\end{figure}

Inspired by minimal criterion co-evolution \cite{MCC}, Wang et al. invented the Paired Open-Ended Trailblazer (POET) \cite{POET}. In POET, environments evolve open-endedly, while agents are optimised to solve them. A minimal criterion ensures that the environments are appropriately difficult for the agents, increasing in complexity as the agents learn more efficient behaviours. Wang et al. show that the environments are used as stepping stones, enabling the agents to learn new skills, and escape local optima. We modify the part of the algorithm that optimises the agents within their environments, in order to allow POET to modify the agents' morphologies as well. The flow of POET, and the genetic algorithm we use, can be seen in fig. \ref{fig:poet_ga_loop}. We explore whether the effect that enabled the controllers evolved with POET to escape local optima, can reduce the problem of premature convergence of morphologies.

\begin{figure}[h!]
\centering
\includegraphics[width=\textwidth]{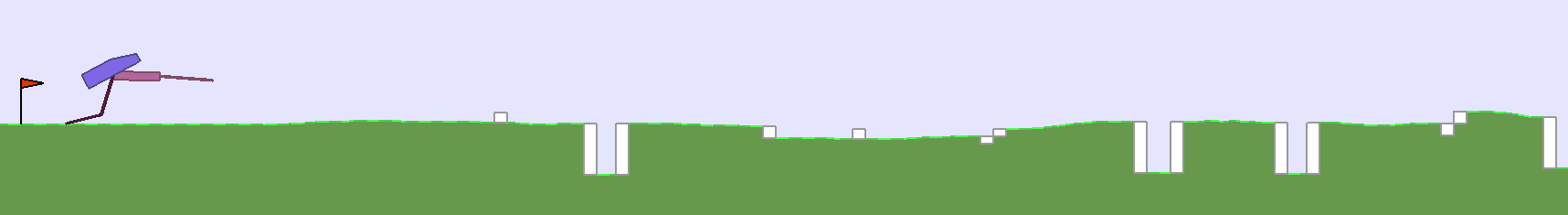}
\caption{The Bipedal Walker Hardcore environment, adjusted to allow for morphological changes.}
\label{BW_figure}
\end{figure}

Our agents are tested in the OpenAI Gym environment Bipedal Walker \cite{bipedal_walker}, Fig. \ref{BW_figure}. We compare the performance and diversity of the agents evolved with POET to agents evolved in two handcrafted curricula of environments. There are two main contributions in this paper: 1) We show that the environments and algorithm structure of POET encourages morphological diversity, and 2) We show that the agents evolved in our handcrafted curriculum with rapid environmental change generalise well to many new environments, while the agents evolved in POET generalise to environments somewhat similar to the ones the agents have seen previously.

\section{Background} 

In 1994 Karl Sims published a study, "Evolving Virtual Creatures" \cite{KarlSims}, which showed virtual creatures evolving in an artificial world with simulated physics. In this artificial evolution the creatures evolved both their bodies and their behaviours simultaneously, and solved various tasks such as walking, swimming and competing against each other. Inspired by Sims' work, many researchers took interest in creating robots that can evolve both their morphology and controller at the same time \cite{RLforAD, diff, scale, ALPS, lama}, and the field of co-optimisation of robot morphology and controller emerged.
Although computing power has increased significantly since Karl Sims' study was published, the morphological complexity of evolved agents has not increased as much as could be expected \cite{GeijtenbeekT.2012ICAU}. Deficiencies in the morphology encodings \cite{HornbyG.S2001Eogd}, deficiencies in the diversity maintenance of search algorithms \cite{NSLC}, and that the environments used are not complex enough to encourage complex morphologies \cite{Auerbach}, have been suggested as sources for this problem.
In 2012 Cheney et al.\cite{diff} proposed their theory on why it was so difficult to make further progress. When co-optimising morphology and controller, the morphology would often converge prematurely. Cheney et al. proposed that this is due to an effect called embodied cognition. 

Embodied cognition is a theory stating that how a creature behaves is influenced heavily by their body. The body can be seen as an interface between the controller and the environment, and if the body changes, even just a little, it is as if the interface between body and controller has been scrambled. The controller adapts to its specific morphology, and when the morphology changes, the controller will have to re-adapt before it can manage to locomote with the new body.
Cheney et al. \cite{scale} continued their research, and studied how explicitly protecting individuals that had just experienced morphological change affected the evolution of morphologies. They showed that when giving the controllers time to adapt to their new bodies, controllers that normally would have been discarded due to low fitness were kept, as they surpassed the previous elites during the time they were protected.

Several algorithms that reduce the problem of embodied cognition, without explicitly protecting novel morphologies, have also been proposed. One such algorithm is ALPS \cite{ALPS}, where reproduction is only allowed between candidates that have experienced approximately the same number of earlier reproduction steps. This restriction divides the population into layers based on their age, and lowers the selection pressure on young candidates.
Jelisavcic et al. \cite{lama}, also take a more indirect approach to protecting new morphologies. In their work all controllers adapt to their morphologies, before being evaluated, through lamarckian evolution. 
Lehman et al. \cite{NSLC} do not allow the controllers time to adapt to their morphologies, but rather increase morphological diversity by optimising for morphological novelty in addition to performance with a multi-objective evolutionary algorithm.

\subsection{POET}

In traditional evolutionary algorithms it is common to optimise for better performance, but this approach can easily lead the algorithm to converge to a local optimum prematurely. One way to increase the chance of finding good optima is to increase diversity in the population, with methods such as fitness sharing \cite{fitshare1, goldberg1987genetic}, speciation \cite{speciation} or crowding \cite{crowding}. However, open-ended algorithms such as novelty search \cite{NS} have also proven efficient. In the field of open-endedness, the focus is not to move towards solutions with better performance, but to create novel and interesting solutions \cite{OverviewOE}, often by optimising for diversity instead of performance. Counterintuitively, searching for novelty alone can sometimes lead to better solutions than what can be found by optimising directly for performance, as demonstrated by Lehman et al. \cite{NS}.

The Paired Open-Ended Trailblazer, POET, is an open-ended algorithm created by Wang et al. \cite{POET}. POET has a population of pairs, where each pair consists of one environment and one agent. The agents are optimized within their paired environment, and the environments are evolved with an open-ended algorithm optimising for novelty. 
As the environments increase in complexity, the agents learn increasingly complex behaviours. Wang et al. tested their algorithm in the OpenAI bipedal walker environment \cite{bipedal_walker}, and observed that the agents used the environments as stepping stones to learn behaviours and gaits they would otherwise not find. The pairs share their knowledge through agent transfers, helping each other escape local optima.

The POET algorithm starts by initialising one environment-agent pair. This first pair always has a very simple environment, such as flat ground. The main flow of the POET algorithm has three steps: Creating environments, Optimising agents in their paired environments and Transferring agents between environments. For the optimisation step Wang et al. use an evolutionary strategy. The two remaining steps are described in detail below.

\subsubsection{Creating environments}

The environment creation step of POET is executed periodically, with a set number of generations between each execution. This step starts off by checking all pairs against a minimal criterion for reproduction. All environments that have a paired agent with fitness higher than the minimal criterion are marked as eligible to reproduce. If there are no eligible environments the creation of environments is skipped. The new environments are then generated by randomly selecting and mutating qualified environments. 

The newly created environments then need to be assigned an agent to become a pair. All agents are tested in the new environments, and the environments are assigned a copy of the agent that performed best in them. The new pairs are then checked against a second minimal criterion, the minimal criterion of difficulty. This minimal criterion has an upper and lower boundary for agent fitness, and ensures the environment is not too difficult nor too easy for its agent. The new pairs that do not meet this minimal criterion are removed.

The remaining new pairs are then sorted by environment novelty. The novelty of the environment is found by comparing it to an archive of all environments that have existed throughout the run. The novelty measure is the euclidean distance to the five nearest neighbours in the archive. If a child environment already exists in the archive, it is removed from the list of child pairs. The most novel child pairs are added to the population until the maximum number of children that can be added each generation is reached, or until there are no more children left to add. The POET population has a maximum population size. When the population size exceeds this limit the oldest pairs are removed.

\subsubsection{Transferring agents between environments}
\label{section:poet_transfer_agents}

In the transfer step, all agents are cross tested in all environments. 
If any of the agents performs better in an environment than the environment's paired agent, the paired agent is removed, and is replaced by a copy of the agent that performs best.

There are two types of transfer, direct and proposal transfer. In direct transfer the agents are tested directly in the other pairs' environments, while in proposal transfer the agents are first trained in the other pairs' environments before they are tested. Transferring of agents allows skills learned in one environment to be used in another environment, and in this way, the pairs trade experiences.

\section{Methods}
We evolve environment-agent pairs with POET, and compare this approach to evolving agents in hand-crafted curricula of environments. \footnote{Source code can be found at \url{https://github.com/EmmaStensby/poet-morphology}} We have substituted the agent optimisation step of POET with a genetic algorithm. While the optimisation step used by Wang et al. \cite{POET} in POET only evolves the agent controllers, our optimisation step evolves the morphologies of the agents as well, allowing us to look at the effect POET has on morphological development in agents. The flow of the algorithm can be seen in fig. \ref{fig:poet_ga_loop}. In order to reduce the computation time of POET we only use direct transfer, and not proposal transfer. Table \ref{Table:POET_params} summarises the values we use for parameters required by POET.

\begin{table}[h!]
\begin{center}
\begin{tabular}{|l|c|}
\hline
\textbf{Parameter} & \textbf{Value}\\
\hline
Pair population size & 20 \\
Transfer frequency  & Every 5 generations\\
Create env. frequency & Every 40 generations\\
Reproduction criterion & 200\\
Difficulty criterion & 50-300\\
Child environments created & 20\\
Child pairs admitted & 2\\
\hline
\end{tabular}
\end{center}
\caption{POET parameters.}
\label{Table:POET_params}
\end{table}

The environments the agents are evaluated in can contain various features: stumps, pits, rough terrain and stairs. In POET's environment creation step the environments' features are mutated to create new environments. 
The parameters for the environment mutation are summarised in table \ref{Table:ENV_params}.

\begin{table}[h!]
\begin{center}
\begin{tabular}{|l|c|c|c|}
\hline
\textbf{Feature } & \textbf{Minimum Value } & \textbf{Mutation Value } & \textbf{Maximum Value } \\
\hline
Terrain roughness & 0 & uniform(0,0.6) & 10\\
Pit gap & [0,0] & [$\pm$0.4,$\pm$0.4] & [10,10]\\
Stump height & [0,0] & [$\pm$0.2,$\pm$0.2] & [5,5]\\
Stair height & [0,0] & [$\pm$0.2,$\pm$0.2] & [5,5]\\
Stair steps & 0 & $\pm$1 & 9\\
\hline
\end{tabular}
\end{center}
\caption{Environment mutation parameters.}
\label{Table:ENV_params}
\end{table}

\subsection{Genetic Algorithm}
\label{genetic_algorithm}
\begin{figure}[]
\centering
\includegraphics[width=225pt]{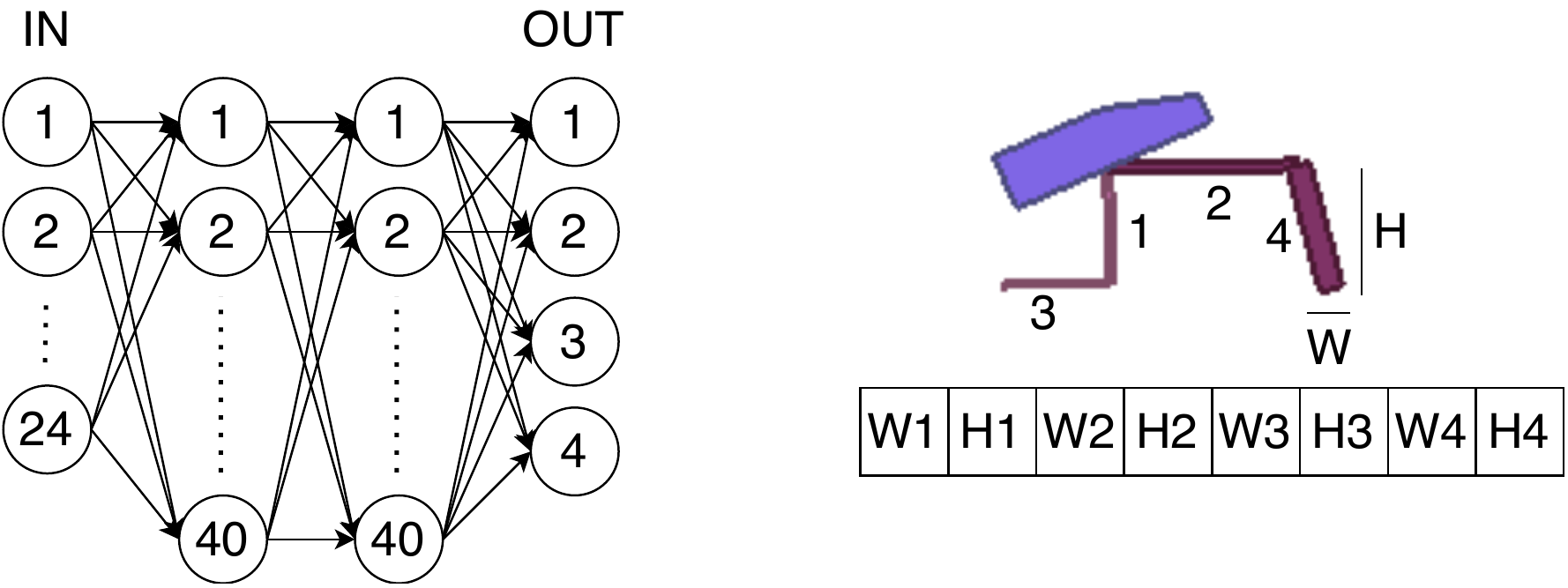}
\caption{Overview of the genotype. The genotype for an individual contains a matrix with the weights of the neural network controller, and a vector with the leg sizes.}
\label{genotype}
\end{figure}

This section describes the genetic algorithm we use to co-optimise controllers and morphologies. The setup and parameters of the genetic algorithm were decided through initial experiments aiming to find values that efficiently evolved high-quality agents. The genetic algorithm keeps a population of 192 individuals, where each individual consists of a neural network and a morphology, see fig. \ref{genotype}. The neural network and morphology controls the behaviour and body of a bipedal walker agent. We will use the term \emph{individual} to refer to a pair consisting of a neural network and a morphology, and the term \emph{agent-population} to refer to a population of 192 individuals being evolved with our genetic algorithm. We use the term agent in agent-population to emphasise that this is the agent part of a POET environment-agent pair. The term \emph{bipedal agent}, is used to refer to the walking figure in the bipedal walker environment.

\subsubsection{Controller}
The bipedal agent is controlled by a neural network inputting state variables such as joint angles, speed, and ground contact sensors, and outputting force to apply to the leg joints, thus forming a type of closed-loop control architecture.  The neural network has an input layer with 24 nodes, two hidden layers with 40 nodes each, and an output layer with four nodes. This gives a total of 2720 weights. The activation function used is the identity function. This network structure has been used in two other studies that also evolved agents locomoting in the bipedal walker environment \cite{RLforAD}\cite{POET}. This design choice was made to reduce the extent of the parameter search. However, it would be interesting to explore whether the same performance could have been reached with a smaller network. The neural network weights are initialised to random values, drawn uniformly between -1 and 1. Mutations can never increase the weights above 30, or decrease them below -30.

\subsubsection{Morphology}
The bipedal walker agent has two legs, each consisting of two segments. The morphology is a vector of eight floats describing the widths and heights of the bipedal agent's four leg segments. The sizes are constrained to values within $\pm 75\%$
of the leg sizes in the original bipedal walker environment. These constraints have previously been used by Ha et al. \cite{RLforAD}. The morphology values are initialised to random values drawn uniformly between the minimum and maximum possible size.

\subsubsection{Individual Evaluation}
\label{individual_evaluation}
To evaluate the fitness of an individual it is tested in a bipedal walker environment. The individual receives positive reward each time step for how far the bipedal agent moved forward, and negative reward for how much force it applied to the joints. It also receives negative reward if the bipedal agent's head touches the ground. The simulation ends when the bipedal agent's head touches the ground, it reaches a flag at the end of the course or 1000 time steps is reached. The reward an individual gets in an environment is not deterministic, and can be unstable. To make the fitness function more stable the individuals are evaluated four times. The fitness is the mean of the reward received in the four evaluations. The negative reward received from applying force to the joints or falling can sometimes exceed the positive reward earned from walking forward, causing individuals to have negative fitness.

\subsubsection{Parent Selection}

The parents are selected by tournament. Five individuals are chosen at random from the population, and compete with their fitness to become a parent. This is repeated until 192 parents have been chosen. The same individual can be chosen as a parent multiple times. The parents are then separated into 96 pairs, and the two parents from each pair are recombined to create two children.

\subsubsection{Recombination and Mutation}

The parents are recombined using uniform crossover. For each neural network weight, or morphology value, the parent contributing the value is chosen at random, with equal probability between the two parents. The first child gets the chosen values, and the second child gets the remaining values.
After recombination the children are mutated using two types of mutation: replacement and modification. In replacement mutation, neural network weights, and morphology values, are chosen with a probability of 0.0075. The chosen values are replaced with new values. The new values are determined in the same way as initial weights and morphology values were determined at individual initialisation.
In modification mutation, neural network weights, and morphology values, are chosen with a probability of 0.075. An offset is added to the chosen weights and values. The offset is a random float drawn uniformly from (-\textit{x}, \textit{x}). For the neural network weights \textit{x} is 0.2. For the morphology values \textit{x} is 16\% of the difference between the minimum and maximum values for the size of the respective leg segment.

\subsubsection{Survivor selection}

To create niches of different solutions in the population, and to slow down convergence, deterministic crowding \cite{crowding} is used when selecting survivors for the next generation.
The difference between two individuals is the L1-norm of the individuals' morphologies. We compare only the morphologies, and not the neural networks, to encourage the niches in the population to explore different morphologies.

\subsection{Environment Curricula}
\label{curricula}
We use two handcrafted curricula of environments in our experiments. The genetic algorithm used to evolve the POET agent-populations is also used to evolve agent-populations in the curricula. The first curriculum is \emph{Static}, which only contains one flat featureless environment. Static is used as a baseline, and has no environmental change. The second curriculum is \emph{Round Robin Incremental} (RRI), this curriculum consists of five environments. The agent-populations are trained for five generations in each of the environments. When training has finished in the last of the five environments, it starts over again at the first environment. The five environments consist of one flat featureless environment, and four environments with features. Each environment has only one feature. The environments with features appear in the following order: Pits, Rough terrain, Stumps, Stairs. The features start out simple. When an individual in the agent-population reaches 150 or higher fitness in any of the last four environments the difficulty of that environment is increased.

\section{Results}
In our experiment we want to find out how effective POET is in creating an environment curriculum that maintains both quality and morphological diversity in a population. 
We evolve agent-populations in the dynamically changing environments of POET, and in the two curricula Static and RRI, with a budget of 384000 evaluations per run. An evaluation being one individual evaluation as described in section \ref{individual_evaluation}. The experiments were performed on a 40 core node on the UNINETT Sigma2 Saga supercomputer, and each run took about 1895 cpu hours to finish. Ten runs are performed in each of the Static, RRI and POET setups. If POET is effective in maintaining both morphological diversity and quality at the same time, we expect the algorithm to find high fitness solutions for many different morphologies.

\subsection{Morphological differences}
\begin{figure}[]
\includegraphics[width=240pt]{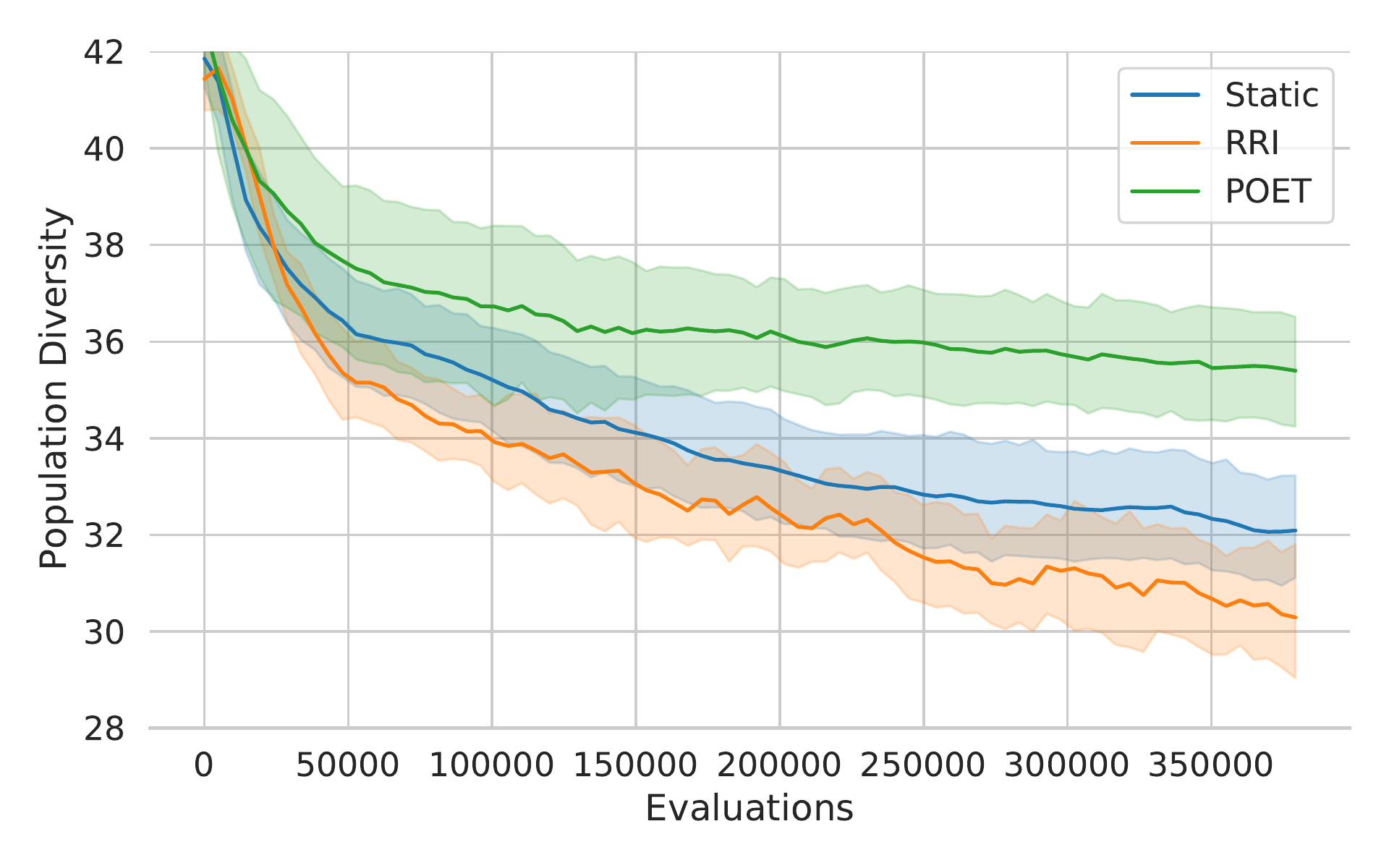}
\includegraphics[width=100pt]{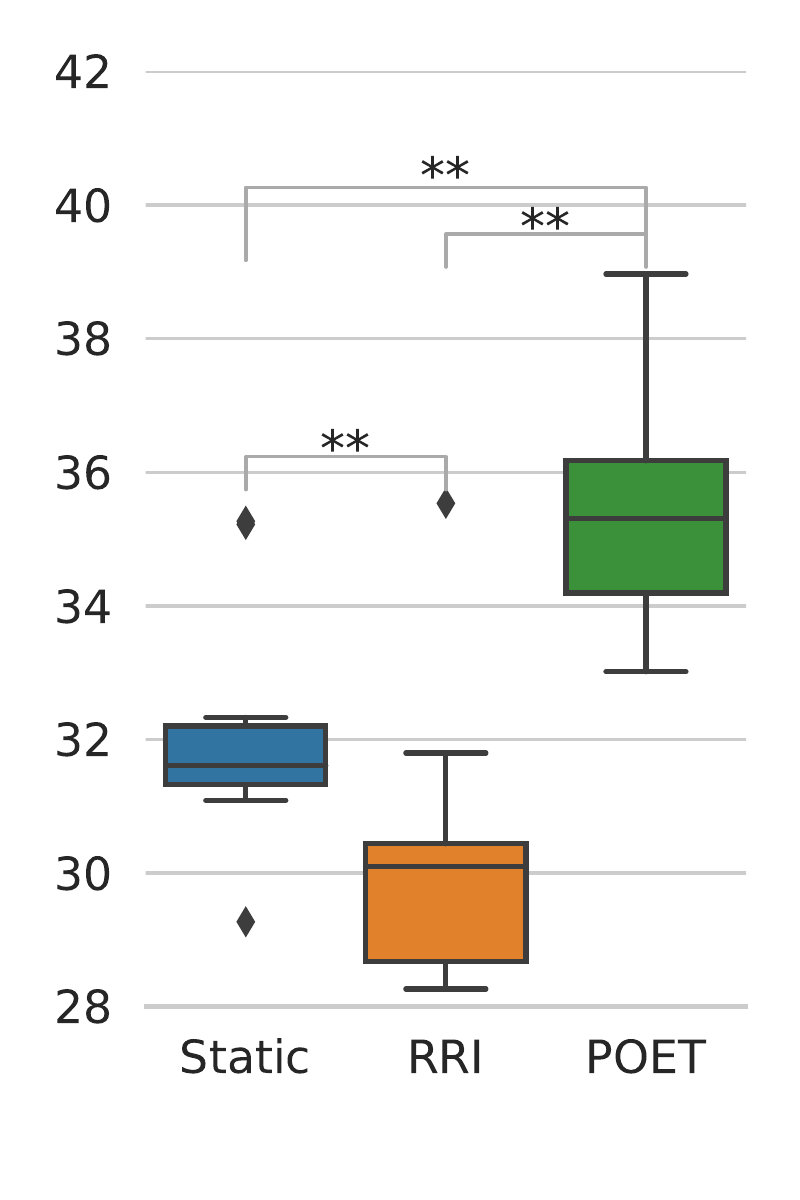}
\caption{\textbf{Left:} Morphological diversity through evolution. The graphs shows the mean of five runs, and the scratched area shows the standard deviation. \textbf{Right:} Morphological diversity in the final populations. Four to one asterisks indicate respectively p $<$ 0.0001, p $<$ 0.001, p $<$ 0.01 and p $<$ 0.05 (Mann-Whitney U test with Bonferroni correction).}
\label{morphological_change}
\end{figure}
Fig. \ref{morphological_change} shows the morphological diversity throughout the runs, for agent-populations evolved in Static, RRI and POET, as well as the morphological diversity of the populations in the final generation. The population diversity of POET in these graphs is measured only for the first POET pair, meaning that we follow an agent-population evolving in a flat environment. However, the POET agent is sometimes switched due to agent transfers, see section \ref{section:poet_transfer_agents}. The morphological diversity of an individual is measured as the average distance from that individual to the other individuals in the population, and the population diversity of an agent-population is the average diversity of all the individuals in the population. POET has higher population diversity than Static and RRI throughout the whole run, and in the resulting populations at the end of the runs (p $<$ 0.01).

\begin{figure}[]
\includegraphics[width=170pt]{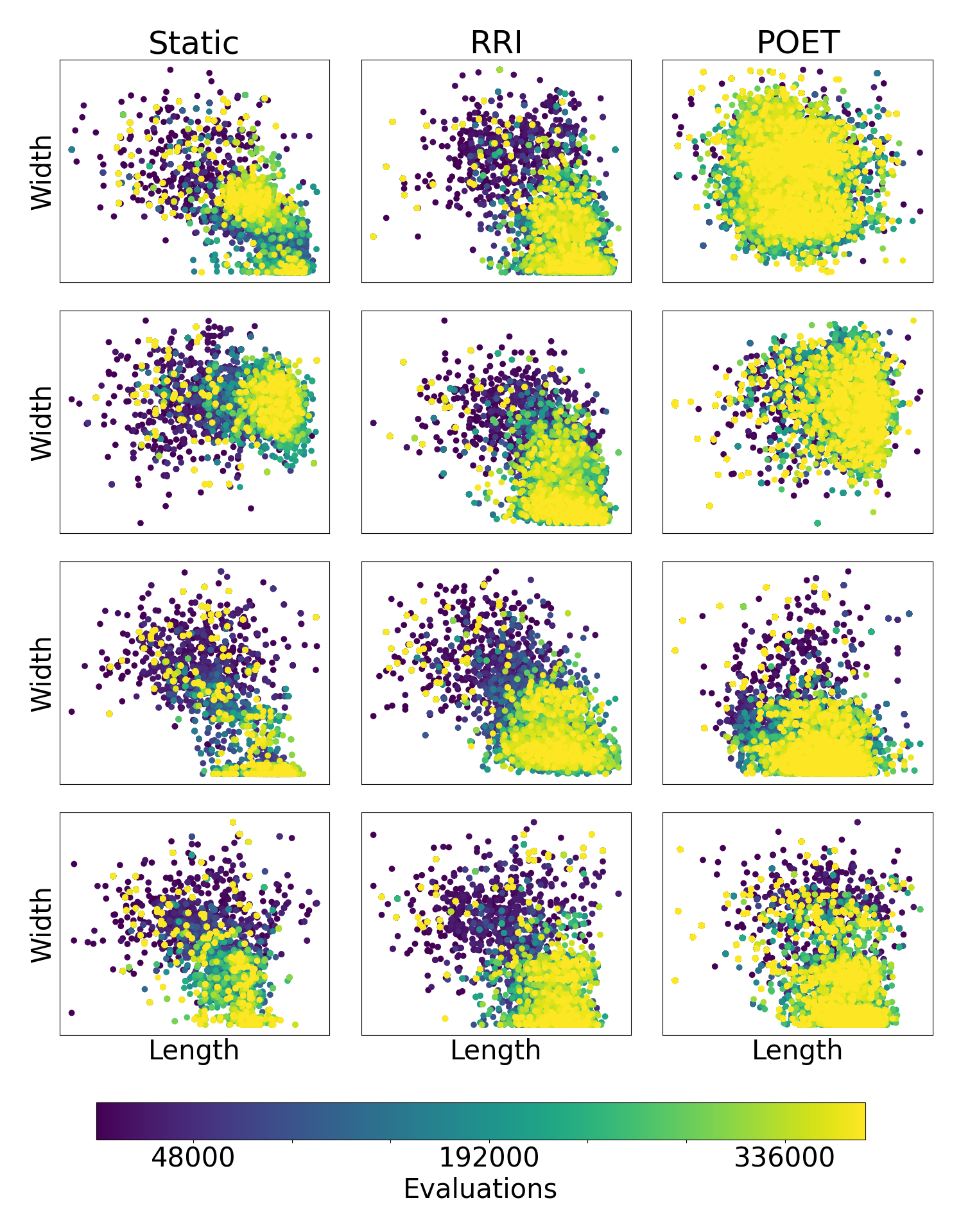}
\includegraphics[width=170pt]{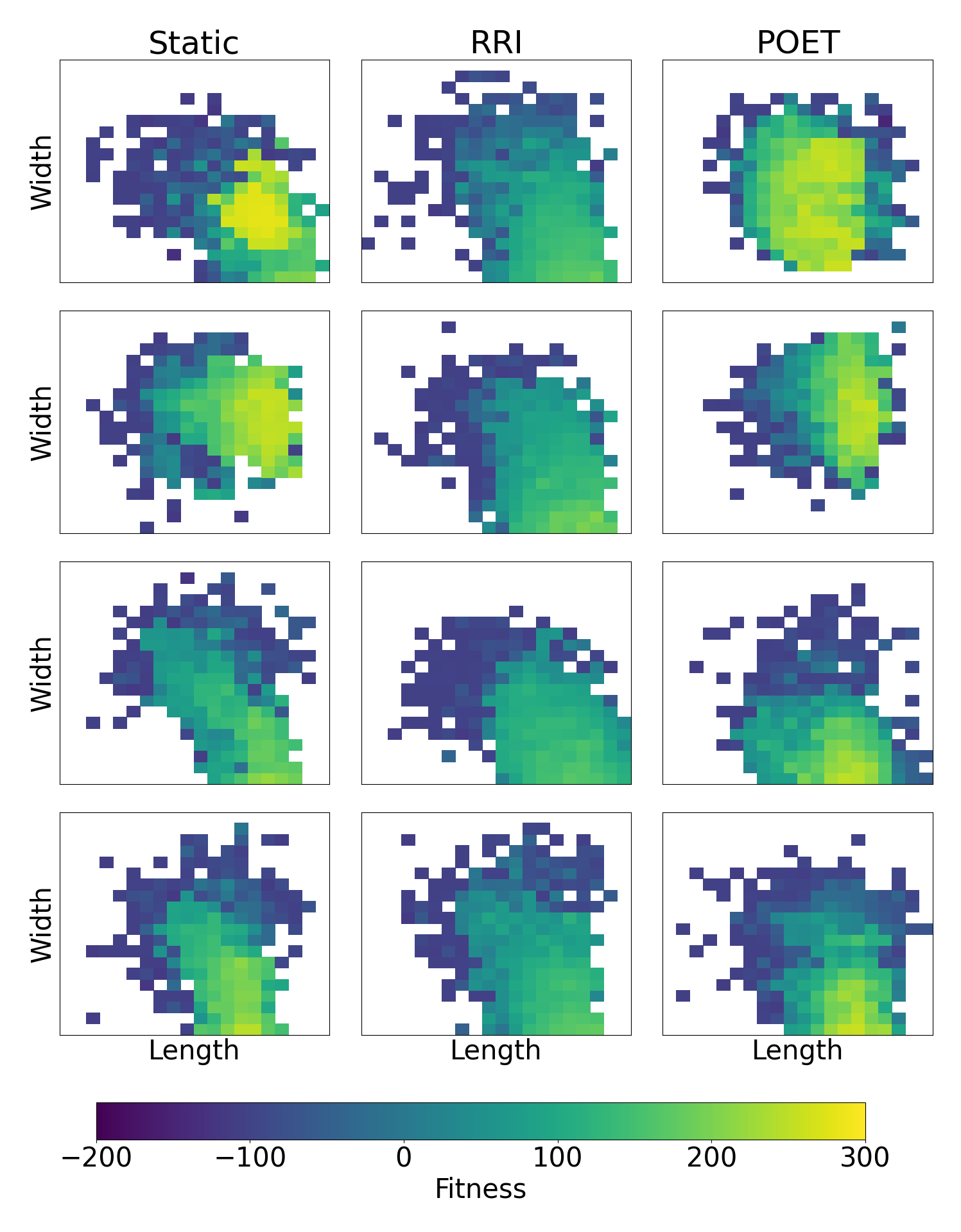}

\caption{Each map is from a separate run of one of the algorithms, and shows the total width and the total length of the legs of all individuals encountered throughout the run. \textbf{Left:} Morphological feature maps. Every morphology that appears throughout the run is represented as a circle. The color of the circle represents when in the search the morphology appeared. \textbf{Right:} Quality-Diversity feature maps. The feature space is divided into a grid, and the color of each cell represents the highest fitness found in that region throughout the run.}
\label{feature_maps}
\end{figure}
Fig. \ref{feature_maps} shows morphological feature maps and quality-diversity feature maps for Static, RRI and POET. To create the feature maps we project the morphological search space into two dimensions. In our case the dimensions are the total length and width of the bipedal agents' legs.
In the morphological feature maps we can see that for RRI and Static the morphologies tend to start out in one area and then collect in a smaller section in the lower right of the search space, while POET covers more of the same areas at the beginning and the end of the runs.

In the quality-diversity feature maps, especially in the run on the top row, we can see that Static has found very high fitness for a few morphologies, while POET has found good controllers for a larger section of the feature space. RRI has the lowest fitness. 
However, it has found mediocre solutions (fitness values around 100) for large sections of the feature space. 

\subsection{Robustness of solutions}

\begin{figure}
\includegraphics[width=340pt]{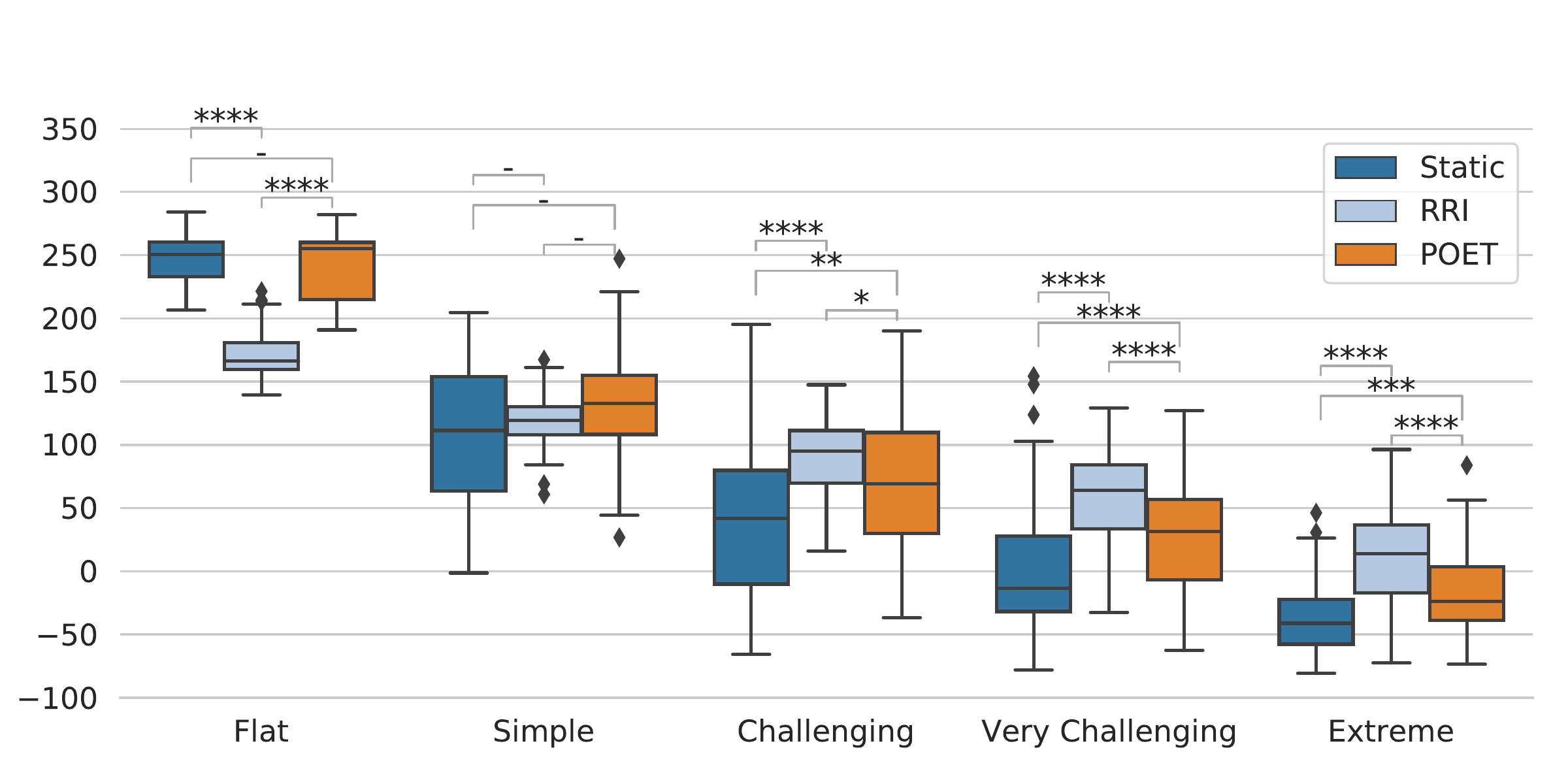}
\caption{The agent-populations tested in randomly selected unseen environments of increasing difficulty. Four to one asterisks indicate respectively p $<$ 0.0001, p $<$ 0.001, p $<$ 0.01 and p $<$ 0.05 (Mann-Whitney U test with Bonferroni correction).}
\label{generalisation_fig}
\end{figure}

Next we tested the robustness of the agent-populations by looking at their performance in environments they had not seen during training. This is shown in fig. \ref{generalisation_fig}. We divided the environmental search space into five categories based on difficulty. The environment difficulty is based on the definition used by Wang et al. \cite{POET}. The first category has flat featureless environments. The environments in the simple category have three features: stumps, pits and terrain roughness. However, the sizes of all the features are below difficulty thresholds. The thresholds are: 3.0 for terrain roughness, 2.5 for pits, 1.5 for stumps.
The environments in the next three categories have respectively one, two or three features with values above its threshold. 10 environments were generated for each category, giving a total of 50 environments. We can see that POET and Static perform better than RRI in the flat category (p $<$ 0.0001), and all three perform similarly in the simple category. However, as the difficulty of the environments increase RRI seems more robust, as it has a lower fitness loss compared to POET and Static, and performs best in the most difficult category (p $<$ 0.0001).

\begin{figure*}[h!]
\includegraphics[width=340pt]{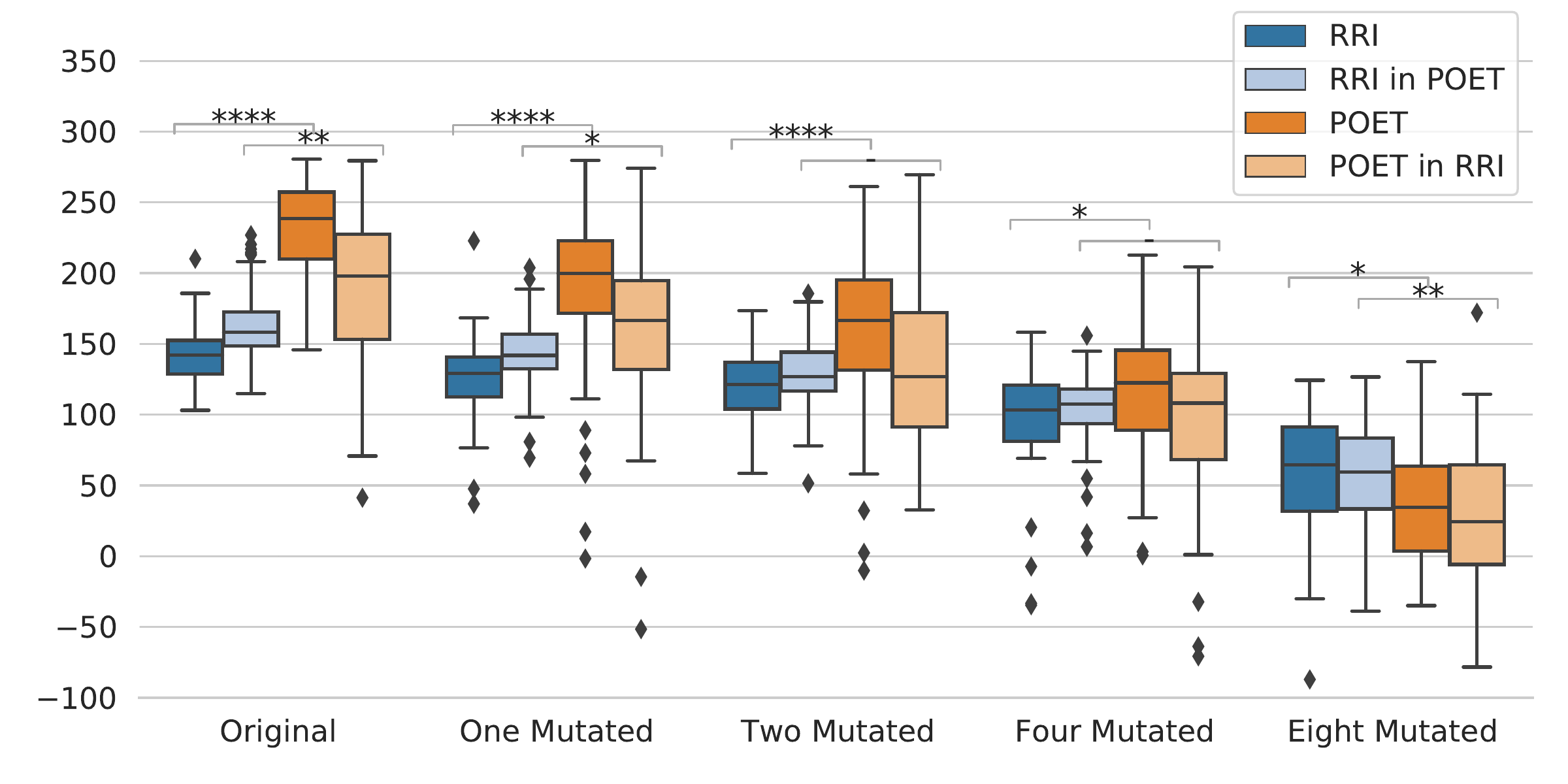}
\caption{The agent-populations tested in randomly selected environments from the vicinity of the agent-populations' original environments, with increasing difficulty. Four to one asterisks indicate respectively p $<$ 0.0001, p $<$ 0.001, p $<$ 0.01 and p $<$ 0.05 (Mann-Whitney U test with Bonferroni correction).}
\label{local_generalisation_fig}
\end{figure*}

In fig. \ref{local_generalisation_fig} we observe the agent-populations' generalisation to environments similar, but slightly different, to the ones seen during training. Here the first environment class consists of the original environments that the agent-populations encountered during training. In the following classes the environments have been mutated, respectively one, two, four or eight times, to increase their difficulty. A mutation is done by choosing a random feature, and adding an offset to the feature variable. The offset is 2.4 for terrain roughness, and 0.8 for the other features. We can see that in the first and second classes, where the environments are very similar to POET and RRI's original environments, POET performs best (p $<$ 0.05). However, as the difficulty of the environments increase POET loses fitness faster than RRI, and in the last class RRI performs better than POET(p $<$ 0.05). 

\section{Discussion}

POET seems to explore approximately the same areas of the morphological feature space in the beginning and end of the evolution, as we saw in fig. \ref{feature_maps}. This suggests that morphologies that are not easy to exploit are kept by POET even if the algorithm does not quickly find a good controller for it. From the feature maps it looks like the Static and RRI agent-populations quite quickly discard morphologies that do not have a good controller, leading to the large purple sections in the morphological feature maps. These maps, together with the high morphological diversity observed for POET in fig. \ref{morphological_change}, lead us to believe that the POET agent-populations have slower morphological convergence than the Static and RRI agent-populations. 

\begin{figure}
\includegraphics[width=\textwidth]{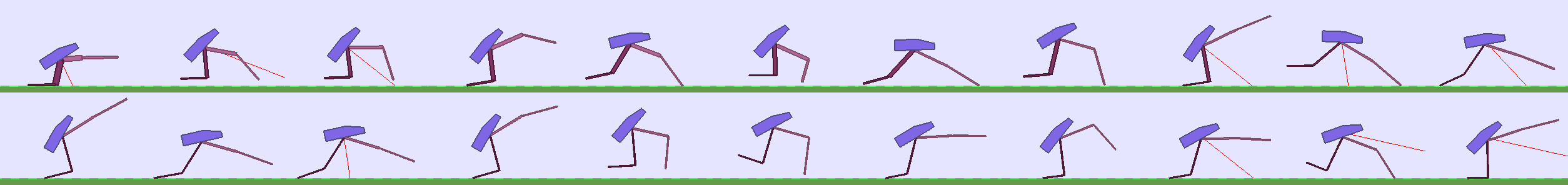}\\
\includegraphics[width=\textwidth]{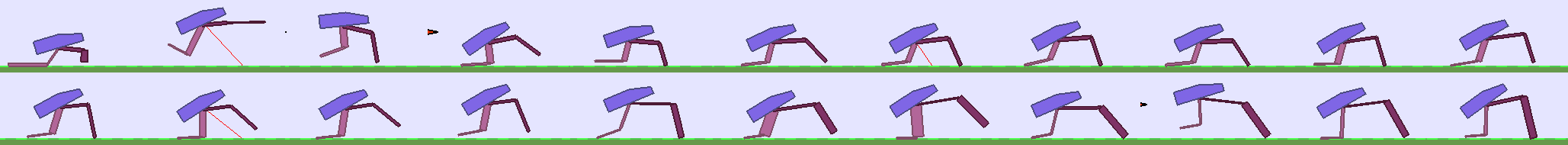}
\caption{The top performing morphology in the flat environment captured every 40 generations for 880 generations. \textbf{Top two rows:} Morphologies from the Static curriculum. \textbf{Bottom two rows:} POET morphologies.\label{morphology_example}}
\end{figure}

We think this may be caused partly by the environmental change, as the RRI agent-populations also have slightly larger yellow areas in the feature maps compared to the Static agent-populations, but also by POET's population size gradually increasing as new environment-agent pairs are added. The pairs can act as niches, exploring different sections of the search space. An example demonstrating the morphological development in a Static and POET agent-population throughout a run can be seen in fig. \ref{morphology_example}. We see that in this run the Static agent-population converged to long thin legs while the POET agent-populations' legs were still evolving. 

As seen in fig. \ref{generalisation_fig} the RRI agent-populations have lower fitness in the flat environment than the agent-populations from Static and POET.
However, four out of five of the environments in the RRI curriculum increase in difficulty whenever the agent-population reaches 150 fitness. This is likely what causes the RRI agent-populations to usually not reach much more than 150 fitness, even in simple environments. The RRI agent-populations have likely sacrificed gait speed to generalise well to the four difficult environments. The rapidly changing environments in RRI force the agent-populations to take a careful approach, as failing in just one of the five environments will likely mean that an individual is removed from the population. This can also explain the lower diversity observed in RRI. Only the individuals that locomote reasonably well in all five environments have high chances of survival, causing the algorithm to quickly converge to this type of individual. However, the RRI agent-populations were more robust to increase in environmental difficulty, as seen in both fig. \ref{generalisation_fig} and fig. \ref{local_generalisation_fig}, where RRI performed better than the Static and POET agent-populations in the most difficult environments. This suggests that the quick environmental change in RRI encourages generalisation to new environments.

Evolving in multiple environments in parallel should naturally promote a diverse set of morphological strategies.
This can be observed in the quality-diversity maps: we see that POET tends to have larger high-fitness areas compared to RRI and Static, meaning that POET has found high fitness for a larger variety of morphologies.
In figure 4 we saw that, although POET and RRI both differ from static in that they experience several different environments, POET had increased population diversity, while RRI had decreased population diversity. RRI likely had decreased diversity due to the difficult nature of its rapidly changing environments, leading RRI to quickly exploit the solutions that were the most robust. The population diversity for POET in this graph is the diversity within the population of a single pair, so the increased diversity is not due to the pairs acting as niches and thus being different from each other. Rather, we believe that this increased diversity is due to the transfer mechanism in POET. The morphologies that are best for the agent’s current environment may not be best for some other environment. Agent-populations with diverse populations may therefore be more likely to be transferred, as diversity may increase its chances of being a good fit for a new environment. When an agent is transferred it is duplicated, thus increasing its presence in the population. Further experiments would be necessary to confirm this.

\section{Conclusion}

In our experiments we compared agent-populations evolved in a static environment, in a curriculum of environments and in POET. We observed that the agent-populations evolved in POET had higher morphological diversity than the agent-populations evolved in a static environment, or in a curriculum of environments. 
This correlation suggests that evolving agent-populations with POET causes increased exploration of morphologies.
This property could be promising for tackling the challenging problem of stagnation in co-optimisation of morphology and control in robotics, and suggest that POET could be applied to this domain.

We also compared the robustness of agent-populations when they encountered unseen environments. We conclude that the agent-populations evolved in the hand crafted curriculum, RRI, were most robust to environmental change. The agent-populations evolved in RRI performed best in the most difficult environments, but their fitness values were relatively low across all environments.
The POET agent-populations had high fitness values in their original environments, and generalised quite well to environments slightly different from these.

POET requires a lot of time and computation power to reach the most difficult environments. Due to limited time, the environments POET found in our runs were not very difficult. It would have been interesting to see how the diversity, robustness and morphological convergence developed if the algorithms were allowed to run longer.

In future work it would be interesting to to test our approach in a complex domain, where the morphologies could evolve more freely, such as on modular robots. We used the bipedal walker environment in order to be able to compare to previous research using this environment \cite{POET, RLforAD}. However, the differences in morphologies might have been more prominent had we used an environment with more complex morphologies.  It would also be interesting to look at how environment curricula could be created to most efficiently promote both high quality, robust individuals and exploration of morphologies. This could be explored either by looking at how features in hand crafted curricula affect the search, or by attempting to create an algorithm similar to POET that is more computationally efficient, and encourages more frequent environmental change. Perhaps this could be achieved by using information about the progress of the search, or the morphological diversity of the agent-populations, when choosing the next environment. We would also like to compare this approach to other approaches that increase morphological diversity, such as protection of individuals that have experienced morphological change \cite{scale}, NSLC \cite{NSLC} or map-elites\cite{nordmoen2020map}.

\section*{Acknowledgments}
This work was partially supported by the Research Council of Norway through its Centres of Excellence scheme, project number 262762.
The simulations were performed on resources provided by UNINETT Sigma2 - the National Infrastructure for High Performance Computing and Data Storage in Norway.

\printbibliography

@inproceedings{goldberg1987genetic,
  title={Genetic algorithms with sharing for multimodal function optimization},
  author={Goldberg, David E. and Richardson, Jon},
  booktitle={Genetic algorithms and their applications: proceedings of the second International Conference on Genetic Algorithms},
  year={1987}
}

@article{speciation,
issn = {0921-0296},
journal = {Journal of Intelligent and Robotic Systems},
pages = {323--351},
volume = {64},
publisher = {Springer Netherlands},
number = {3-4},
year = {2011},
title = {Speciation in Behavioral Space for Evolutionary Robotics},
author = {Trujillo, Leonardo and Olague, Gustavo and Lutton, Evelyne and Fernández de Vega, Francisco and Dozal, León and Clemente, Eddie},
}

@article{fitshare1,
issn = {0304-3975},
journal = {Theoretical Computer Science},
pages = {53--70},
volume = {773},
publisher = {Elsevier B.V},
year = {2019},
title = {On the benefits and risks of using fitness sharing for multimodal optimisation},
author = {Oliveto, Pietro S and Sudholt, Dirk and Zarges, Christine},
}

@article{GeijtenbeekT.2012ICAU,
issn = {0167-7055},
journal = {Computer Graphics Forum},
pages = {2492--2515},
volume = {31},
publisher = {Blackwell Publishing Ltd},
number = {8},
year = {2012},
title = {Interactive Character Animation Using Simulated Physics: A State‐of‐the‐Art Review},
author = {Geijtenbeek, Thomas and Pronost, Nicolas},
}

@article{miras2020environmental,
  title={Environmental influences on evolvable robots},
  author={Miras, Karine and Ferrante, Eliseo and Eiben, Agoston E.},
  journal={PloS one},
  volume={15},
  number={5},
  pages={e0233848},
  year={2020},
  publisher={Public Library of Science San Francisco, CA USA}
}

@article{Auerbach,
      title = {Environmental Influence on the Evolution of Morphological  Complexity in Machines},
      author = {Auerbach, Joshua E. and Bongard, Josh C.},
      publisher = {Public Library Science},
      journal = {PLoS Computational Biology},
      number = {1},
      volume = {10},
      pages = {17. e1003399},
      year = {2014},
      url = {http://infoscience.epfl.ch/record/195214},
      doi = {10.1371/journal.pcbi.1003399},
}

@inproceedings{HornbyG.S2001Eogd,
issn = {10504729},
pages = {4146--4151 vol.4},
volume = {4},
publisher = {IEEE},
booktitle = {Proceedings 2001 ICRA. IEEE International Conference on Robotics and Automation},
isbn = {0780365763},
year = {2001},
title = {Evolution of generative design systems for modular physical robots},
author = {Hornby, Gregory S. and Lipson, Hod and Pollack, Jordan B.},
}

@inproceedings{dyret3,
series = {GECCO '18},
pages = {125--132},
publisher = {ACM},
booktitle = {Proceedings of the Genetic and Evolutionary Computation Conference},
isbn = {9781450356183},
year = {2018},
title = {Real-world evolution adapts robot morphology and control to hardware limitations},
author = {Nygaard, T{\o}nnes and Martin, Charles and Samuelsen, Eivind and Torresen, Jim and Glette, Kyrre},
}

@inproceedings{ALPS,
author = {Hornby, Gregory S.},
title = {ALPS: The Age-Layered Population Structure for Reducing the Problem of Premature Convergence},
year = {2006},
isbn = {1595931864},
publisher = {Association for Computing Machinery},
url = {https://doi-org.ezproxy.uio.no/10.1145/1143997.1144142},
doi = {10.1145/1143997.1144142},
booktitle = {Proceedings of the 8th Annual Conference on Genetic and Evolutionary Computation},
pages = {815–822},
numpages = {8},
series = {GECCO ’06}
}

@inproceedings{MCC,
author = {Brant, Jonathan C. and Stanley, Kenneth O.},
title = {Minimal Criterion Coevolution: A New Approach to Open-Ended Search},
year = {2017},
isbn = {9781450349208},
publisher = {Association for Computing Machinery},
url = {https://doi-org.ezproxy.uio.no/10.1145/3071178.3071186},
doi = {10.1145/3071178.3071186},
booktitle = {Proceedings of the Genetic and Evolutionary Computation Conference},
pages = {67–74},
numpages = {8},
series = {GECCO ’17}
}

@article{crowding,
author = {Mengshoel, Ole J. and Goldberg, David E.},
title = {The Crowding Approach to Niching in Genetic Algorithms},
journal = {Evolutionary Computation},
volume = {16},
number = {3},
pages = {315-354},
year = {2008},
doi = {10.1162/evco.2008.16.3.315},

URL = { 
        https://doi.org/10.1162/evco.2008.16.3.315
    
},
eprint = { 
        https://doi.org/10.1162/evco.2008.16.3.315
    
}
}

@inproceedings{POET,
author = {Wang, Rui and Lehman, Joel and Clune, Jeff and Stanley, Kenneth O.},
title = {POET: Open-Ended Coevolution of Environments and Their Optimized Solutions},
year = {2019},
isbn = {9781450361118},
publisher = {Association for Computing Machinery},
url = {https://doi.org/10.1145/3321707.3321799},
doi = {10.1145/3321707.3321799},
booktitle = {Proceedings of the Genetic and Evolutionary Computation Conference},
pages = {142–151},
numpages = {10},
series = {GECCO ’19}
}

@ARTICLE{lama,
  
AUTHOR={Jelisavcic, Milan and Glette, Kyrre and Haasdijk, Evert and Eiben, Agoston E.},   
	 
TITLE={Lamarckian Evolution of Simulated Modular Robots},      
	
JOURNAL={Frontiers in Robotics and AI},      
	
VOLUME={6},      

PAGES={9},     
	
YEAR={2019},      
	  
URL={https://www.frontiersin.org/article/10.3389/frobt.2019.00009},       
	
DOI={10.3389/frobt.2019.00009},      
	
ISSN={2296-9144}
}

@article{diff,
author = {Cheney, Nicholas and Bongard, Josh and Sunspiral, Vytas and Lipson, Hod},
title = {On the Difficulty of Co-Optimizing Morphology and Control in Evolved Virtual Creatures},
journal = {Artificial Life Conference Proceedings},
volume = {},
number = {28},
pages = {226-233},
year = {2016},
doi = {10.1162/978-0-262-33936-0-ch042},

URL = { 
        https://www.mitpressjournals.org/doi/abs/10.1162/978-0-262-33936-0-ch042
    
},
eprint = { 
        https://www.mitpressjournals.org/doi/pdf/10.1162/978-0-262-33936-0-ch042
    
}
}

@article{scale,
issn = {17425689},
journal = {Journal of the Royal Society, Interface},
volume = {15},
number = {143},
year = {2018},
title = {Scalable co-optimization of morphology and control in embodied machines},
author = {Cheney, Nicholas and Bongard, Josh and Sunspiral, Vytas and Lipson, Hod},
}

@misc{bipedal_walker,
  Author = {Klimov Oleg},
  title = {BipedalWalkerHardcore-v2. https://gym.openai.com},
  Year = {2016},
}

@article{NS,
 author = {Lehman, Joel and Stanley, Kenneth O.},
 title = {Abandoning Objectives: Evolution Through the Search for Novelty Alone},
 journal = {Evol. Comput.},
 issue_date = {Summer 2011},
 volume = {19},
 number = {2},
 month = jun,
 year = {2011},
 issn = {1063-6560},
 pages = {189--223},
 numpages = {35},
 url = {http://dx.doi.org/10.1162/EVCO_a_00025},
 doi = {10.1162/EVCO_a_00025},
 acmid = {2000553},
 publisher = {MIT Press},
}

@article{OverviewOE,
issn = {1064-5462},
journal = {Artificial Life},
pages = {93--103},
volume = {25},
publisher = {MIT Press},
number = {2},
year = {2019},
author = {
Packard, Norman and Bedau, Mark A. and Channon, Alastair and Ikegami, Takashi and Rasmussen, Steen and Stanley, Kenneth O. and Taylor, Tim},
title = {An Overview of Open-Ended Evolution: Editorial Introduction to the Open-Ended Evolution II Special Issue},
}

@inproceedings{KarlSims,
series = {SIGGRAPH '94},
pages = {15--22},
publisher = {ACM},
booktitle = {Proceedings of the 21st annual conference on computer graphics and interactive techniques},
isbn = {0897916670},
year = {1994},
title = {Evolving virtual creatures},
author = {Sims, Karl},
}

@article{RLforAD,
issn = {1064-5462},
journal = {Artificial Life},
pages = {352--365},
volume = {25},
publisher = {MIT Press},
number = {4},
year = {2019},
author = {Ha, David},
title = {Reinforcement Learning for Improving Agent Design},
}

@inproceedings{NSLC,
author = {Lehman, Joel and Stanley, Kenneth O.},
title = {Evolving a Diversity of Virtual Creatures through Novelty Search and Local Competition},
year = {2011},
isbn = {9781450305570},
publisher = {Association for Computing Machinery},
url = {https://doi-org.ezproxy.uio.no/10.1145/2001576.2001606},
doi = {10.1145/2001576.2001606},
booktitle = {Proceedings of the 13th Annual Conference on Genetic and Evolutionary Computation},
pages = {211–218},
numpages = {8},
series = {GECCO ’11}
}

@article{nordmoen2020map,
  title={MAP-Elites enables Powerful Stepping Stones and Diversity for Modular Robotics},
  author={Nordmoen, J{\o}rgen and Veenstra, Frank and Ellefsen, Kai Olav and Glette, Kyrre},
  journal={arXiv preprint arXiv:2012.04375},
  year={2020}
}
\end{document}